\documentclass[sigconf,screen]{acmart}
\usepackage{etoolbox}
\makeatletter
\patchcmd{\NAT@test}{\else \NAT@nm }{\else \NAT@nmfmt{\NAT@nm}}{}{}
\makeatother

\copyrightyear{2024}
\acmYear{2024}
\setcopyright{rightsretained}
\acmConference[FAccT '24]{The 2024 ACM Conference on Fairness, Accountability, and Transparency}{June 3--6, 2024}{Rio de Janeiro, Brazil}
\acmBooktitle{The 2024 ACM Conference on Fairness, Accountability, and Transparency (FAccT '24), June 3--6, 2024, Rio de Janeiro, Brazil}\acmDOI{10.1145/3630106.3659025}
\acmISBN{979-8-4007-0450-5/24/06}

\usepackage{tikz}
\usetikzlibrary{calc}

\usepackage{breakurl}

\begin{document}

\title{The Neutrality Fallacy: When Algorithmic Fairness Interventions are (Not) Positive Action}

\author{Hilde Weerts}
\affiliation{%
  \institution{Eindhoven University of Technology}
  \country{The Netherlands}}
\email{h.j.p.weerts@tue.nl}

\author{Rapha\"{e}le Xenidis}
\affiliation{%
  \institution{Sciences Po Law School}
  \country{France}}
\email{raphaele.xenidis@sciencespo.fr}

\author{Fabien Tarissan}
\affiliation{%
  \institution{Université Paris-Saclay, CNRS, ISP, ENS Paris-Saclay}
  \country{France}}
\email{fabien.tarissan@ens-paris-saclay.fr}

\author{Henrik Palmer Olsen}
\affiliation{%
  \institution{University of Copenhagen \& IEA-Paris}
  \country{Denmark}}
\email{henrik@jur.ku.dk}

\author{Mykola Pechenizkiy}
\affiliation{%
  \institution{Eindhoven University of Technology}
  \country{The Netherlands}}
\email{m.pechenizkiy@tue.nl}

\renewcommand{\shortauthors}{Weerts et al.}

\begin{abstract}
Various metrics and interventions have been developed to identify and mitigate unfair outputs of machine learning systems. While individuals and organizations have an obligation to avoid discrimination, the use of fairness-aware machine learning interventions has also been described as amounting to `algorithmic positive action’ under European Union (EU) non-discrimination law. As the Court of Justice of the European Union has been strict when it comes to assessing the lawfulness of positive action, this would impose a significant legal burden on those wishing to implement fair-ml interventions. In this paper, we propose that algorithmic fairness interventions often should be interpreted as a means to prevent discrimination, rather than a measure of positive action. Specifically, we suggest that this category mistake can often be attributed to \emph{neutrality fallacies}: faulty assumptions regarding the neutrality of (fairness-aware) algorithmic decision-making. Our findings raise the question of whether a negative obligation to refrain from discrimination is sufficient in the context of algorithmic decision-making. Consequently, we suggest moving away from a duty to ‘not do harm’ towards a positive obligation to actively ‘do no harm’ as a more adequate framework for algorithmic decision-making and fair ml-interventions.
\end{abstract}





\maketitle

\section{Introduction}
Various metrics and interventions have been developed to identify and mitigate unfair outputs of machine learning systems. Fairness-aware machine learning (fair-ml) interventions range from (by now traditional) algorithms that pose fairness as a constrained optimisation task~\citep[e.g.,][]{kamiran2012data,hardt2016equality} to, more recently, targeted approaches aimed at mitigating specific concerns at the root of fairness-related harms~\citep[e.g.,][]{guerdan2023ground}. While individuals and organizations have an obligation to avoid discrimination, the use of fair-ml interventions could be interpreted as what legal scholars have coined `algorithmic positive action' under EU non-discrimination law \citep{hoch2021discrimination,xenidis2022algorithmic,muller2023fairness} or, equivalently, `algorithmic affirmative action' under US anti-discrimination law~\citep{hacker2018teaching,bent2019algorithmic,ho2020affirmative,salib2022big}.

Positive action describes a wide spectrum of measures that can be taken to redress past discrimination, ranging from outreach and support programmes to quotas for historically disadvantaged protected groups. Many fair-ml interventions (implicitly or explicitly) rely on legally protected group membership to impose equality constraints. At first blush, this points to a violation of the principle of equal treatment, unless framed as a form of justified positive action. Positive action is not uncontroversial -- the Court of Justice of the European Union (CJEU) has been particularly strict when assessing the lawfulness of quotas. In the United States, a recent landmark ruling by the Supreme Court effectively challenged existing practices regarding race-based affirmative action in college admissions.\footnote{\textit{Students for Fair Admissions v. Harvard}, decided June 29, 2023. The ruling may have implications for affirmative action more generally since it contains several principled statements on how the 14th Amendment to the US Constitution should be interpreted. In US law, debates on affirmative action often revolve around the opposition between anticlassification and antisubordination approaches. The latter posits that equality ``cannot be realized under conditions of pervasive social stratification and argue[s] that law should reform institutions and practices that enforce the secondary social status of historically oppressed groups". The first approach, by contrast, posits ``that the government may not classify people either overtly or surreptitiously on the basis of a forbidden category"~\citep{balkin2003american}. Therefore, anti-classification approaches make it difficult to adopt affirmative action measures to redress historical inequalities.} Interpreting a fair-ml intervention as positive action -- rather than a measure to avoid discrimination -- would therefore impose a great legal burden on those wishing to implement them.

However, categorizing a fair-ml intervention as positive action relies on strong assumptions regarding the neutrality of algorithmic decision-making, which is often an oversimplification of reality. For example, developers of the dating app Breeze recently asked the Netherlands Institute for Human Rights whether adjusting their algorithm to ensure that dark-skinned and non-Dutch users are recommended just as frequently as other users would constitute preferential treatment. The Institute judged that this would not be the case. On the contrary: the Instutite stressed that Breeze is obliged to take measures~\citep{breeze2023}. This raises the question of where the obligation to avoid discrimination ends and where positive action starts.

In this paper, we question the extent to which fair-ml interventions should be interpreted as positive action and when they are best interpreted as measures to avoid discrimination. Our contributions are as follows. First, we explore the lawfulness of fair-ml interventions as positive action. Given the Court's strict stance, we find that if fair-ml interventions are assimilated to quotas, they would likely be deemed incompatible with EU law in light of existing case law. Second, we identify three \textit{neutrality fallacies}: assumptions regarding the neutrality of algorithmic decision-making that bring into question whether fair-ml interventions should be interpreted as positive action. Specifically, we argue that fairness interventions aimed at mitigating measurement bias and disparities in predictive performance should not necessarily be considered a form of positive action and that neither fairness-aware nor unconstrained algorithmic decision-making is value-neutral. Third, our findings expose the limitations of the negative obligation to refrain from discrimination encoded in the law. Therefore, we suggest moving away from a duty to ‘not do harm’ towards a positive obligation to actively ‘do no harm’ as a more adequate framework for algorithmic decision-making and fair ml-interventions.

The remainder of this paper is structured as follows. In Section~\ref{sec:positiveaction}, we set out the place of positive action under EU non-discrimination law. In Section~\ref{sec:algorithmicpositiveaction}, we explore the extent to which `algorithmic positive action' is likely to survive legal scrutiny. In Section~\ref{sec:neutralityfallacy}, we unravel the misleading assumption that underpins the legal framing of fair-ml interventions as positive action, namely that algorithmic decision-making is neutral. To this end, we present three neutrality fallacies: ``data is neutral", ``predictive models are neutral", and ``algorithmic decision-making is neutral". In Section~\ref{sec:positiveobligation}, we discuss the limitations of the current negative obligation to avoid discrimination. Section~\ref{sec:conclusion} concludes the paper.  

\section{Positive Action in EU Non-Discrimination Law}
\label{sec:positiveaction}
Positive action is a range of measures that can be taken to redress past discrimination, ranging from outreach measures (e.g., selective advertising) and advancement plans (e.g., vocational training programmes), to preferential treatment for members of historically disadvantaged groups (e.g., quotas). Under EU non-discrimination law, positive action is allowed, but only under very specific circumstances.

\subsection{Positive action in EU legislation}
Theoretical discussions of positive action are generally rooted in the distinction between two conceptions of equality: formal equality and substantive equality. \textit{Formal equality} generally refers to equal treatment, without necessarily referring to protected characteristics. \textit{Substantive equality}, on the other hand, recognizes that to achieve equal outcomes, existing differences between protected groups need to be taken into account. In other words: if one starts from a disadvantaged position due to historical inequalities, formal equal treatment alone is insufficient to achieve equal outcomes. A third label, \textit{transformative equality}, has been coined in the context of the UN Convention on the Elimination of All Forms of Discrimination Against Women (CEDAW) and refers to measures like so-called temporary special measures that aim to transform institutions and practices that perpetuate gender segregation and stereotypes~\citep{cusack2013cedaw}.

These equality models map onto a core distinction in EU non-discrimination law. The ban on \textit{direct discrimination} prohibits less favourable treatment on grounds of a protected characteristic. This means that protected characteristics cannot be explicitly included in decision-making processes covered by EU non-discrimination law to treat people less favourably than others. The direct discrimination doctrine has been said to promote formal equality because it encapsulates the first (symmetrical) part of the Aristotelian principle according to which likes should be treated alike. \textit{Indirect discrimination}, on the other hand, occurs when an "apparently neutral provision, criterion, or practice would put persons of a protected group at a particular disadvantage compared with other persons". However, unequal outcomes can be lawful if the provision, criterion, or practice can be objectively justified by a legitimate aim and passes the so-called proportionality test. The ban on indirect discrimination thus allows addressing proxy discrimination and, to some extent, accounts for an unjust status quo. For example, in industries where a gender pay gap exists, using an employee's past salary to decide on their new salary could result in indirect discrimination on the grounds of sex, because it perpetuates the pay gap.\footnote{This is why Article 5(2) of the new Pay Transparency Directive 2023/970 foresees that "[a]n employer shall not ask applicants about their pay history during their current or previous employment relationships".} Indirect discrimination has therefore been described as promoting a form of substantive equality. It speaks to the second (asymmetrical) part of the Aristotelian principle, according to which those who are unalike should be treated in an unalike manner.

However, issues related to the selection of an appropriate comparator and the possibility of an objective justification limit the ability of the indirect discrimination doctrine to achieve substantive equality. For example, the gender pay gap can be partially explained by issues of gender segregation on the labour market, including the overrepresentation of women in relatively low-paying sectors, such as healthcare and education. While the general undervaluation of such feminised jobs perpetuates the gender pay gap, courts might not consider this to be indirect discrimination because the equal pay test requires comparing female workers to male workers who perform `equal work' or `work of equal value'.\footnote{The choice of comparator is a thorny issue in EU non-discrimination law. See \citet{weerts2023algorithmic}. The case law of the CJEU on equal pay allows comparing male and female workers who perform work that may be different but is of equal value, for example, the work of midwives and clinical technicians or that of ceramic painters and automatic-machine operators provided that their working conditions are determined by a single source, e.g., the same employer. See \citet[][para. 42-43]{C-400/93} and \citet[][ para. 48-50]{C-236/98}.} Similarly, differences in educational attainment across protected groups could serve as an objective justification for employment selection procedures that reproduce these inequalities - even if the differences between groups are at least partially caused by structural inequalities due to past discrimination~\citep{hellman2023big}.

The direct and indirect discrimination doctrines thus arguably introduce an obligation to avoid unequal treatment and, to a limited extent, replication of an unjust status quo~\citep{xenidis2018positive}. As the Court explains in \textit{Kalanke}~\citep{C-450/93}, "existing legal provisions on equal treatment, which are designed to afford rights to individuals, are inadequate for the elimination of all existing inequalities unless [...] action is taken [...] to counteract the prejudicial effects [...] which arise from social attitudes, behaviour, and structures". Positive action measures go a step further towards substantive (or transformative) equality and aim to help members of protected groups overcome existing disadvantages caused by historical discrimination. Such measures will, by definition, require awareness of a protected characteristic, causing tension between asymmetrical treatment through positive action measures and the principle of formal equality understood as symmetrical treatment. In particular, preferential treatment on account of a protected characteristic violates such a conception of equal treatment and could be considered unlawful direct discrimination.

In 1976, the Council addressed this potential barrier to positive action in the Equal Treatment Directive \citep{76/207/EEC}, which included a provision that allowed for positive action in the employment domain. Following Article 2(4) of that (now repealed) Directive, Member States could adopt "measures to promote equal opportunity for men and women, in particular by removing existing inequalities which affect women's opportunities". In 1999, Article 141(4) of the Treaty of Amsterdam (now Article 157(4) TFEU \citep{tfeu}) entered into force and  allowed measures that provide "for specific advantages in order to make it easier for the under-represented sex to pursue a vocational activity or to prevent or compensate for disadvantages" linked to protected grounds. Today, Article 7(1) of the Framework Equality Directive \citep{employmentdirective} and Article 5 of the Race Equality Directive \citep{racialequalitydirective} state that "[w]ith a view to ensuring full equality in practice, the principle of equal treatment shall not prevent any Member State from maintaining or adopting specific measures to prevent or compensate for disadvantages linked to [a protected ground]". Similarly, Article 3 of the Recast Directive \citep{2006/54/EC} entitled "positive action" states that "Member States may maintain or adopt measures [...] with a view to ensuring full equality in practice between men and women in working life". Importantly, EU law does not require -- but only allows -- Member States to adopt positive action measures.\footnote{Except in relation to the rights of persons living with a disability, where Article 5 of the Framework Directive includes an obligation of reasonable accommodation that can be understood as a form of positive action \citep{employmentdirective}.}

\subsection{Positive Action in the Case Law of the CJEU}
Arguably, the prohibition of indirect discrimination entails -- to a certain extent -- an obligation to avoid discrimination and therefore implicitly requires some limited form of positive action~\citep{de2007beyond}. Yet, the Court of Justice has rejected such an interpretation early on in \textit{Bilka-Kaufhaus}~\citep{C-170/84}, explaining that an obligation for an employer "to organize its occupational pension scheme in such a manner as to take into account the fact that family responsibilities prevent women workers from fulfilling the requirements for such a pension [...] goes beyond the scope of Article 119 [now Article 157 TFEU] and has no other basis in Community [now EU] law" \citep[][para. 38, 42]{C-170/84}. In addition, the exact form that lawful positive action can take is not uncontested. While outreach measures are generally accepted, the Court of Justice has been strict when assessing the lawfulness of quotas and has generally treated them as an exception to equal treatment. 

Preference can only be given to members of an underrepresented group if they are otherwise similarly situated and that preference cannot be automatic. For example, in the landmark decision \textit{Kalanke}~\citep{C-450/93}, two applicants, Mr Kalanke and Ms Glissman, were considered equally qualified for the role of section manager. The board decided to give the position to Ms Glissman due to the underrepresentation of women in the corresponding pay bracket. The CJEU ruled this decision unlawful because it gave women "absolute and unconditional priority" and focused on equal results instead of equality of opportunity \citep[][para. 22-23]{C-450/93}. The Court further consolidated this stance in \textit{Marschall}~\citep{{C-409/95}}, where the employer's policy was to grant priority to equally qualified female candidates who apply for a position where women are under-represented. The Court validated the \textit{Marschall} policy because it entailed an objective assessment of the individual situation of all candidates which could reveal "reasons specific to an individual male candidate [and] tilt the balance in his favour" \citep[][para. 24]{C-409/95}. All quota policies must contain such a "saving clause" allowing for exceptions to the quota rule to influence the final decision-making if it is to pass the lawfulness test of the CJEU. In other words, the CJEU has prohibited `rigid' as opposed to `flexible quotas'.
 
The contours of lawful positive action are strict and do not always allow for substantive or transformative equality policies. For example, some policies seeking to correct historical disadvantages upstream (e.g. educational differences between women and men or family-related career interruptions) have not passed legal scrutiny. In \textit{Abrahamsson}, a university policy that permitted the hiring of a sufficiently qualified woman over a better-qualified man was struck down by the Court \citep[][para. 56]{C-407/98}. In addition, the requirement for objectively assessing whether individual candidates with diverse profiles are similarly qualified has been criticised for being impractical and for leaving room for arbitrariness and prejudices.\footnote{In \textit{Abrahamsson} for example, the female applicant's stronger experience in administrative tasks did not count as `a decisive factor' in the qualifications assessment \citep[][para. 23]{C-407/98}. This reflects a broader problem: women are often socialised into taking on tasks that do not count or count less in assessments of professional qualifications and experience.}

The Court has been less strict when it comes to outreach measures or vocational training programs. For example, in \textit{Badeck}~\citep{C-158/97}, the Court allowed ``to provide for a minimum percentage of women, at least equal to the percentage of women among graduates, holders of higher degrees and students in each discipline". Again, however, the Court stressed that automatic preference is not allowed: all candidates must be ``subject to an objective assessment which takes account of the specific personal situations of all candidates", leaving the meaning of "objective" open to interpretation. Moreover, positive action measures should be crafted in a clear and narrow manner and should aim to redress -- as opposed to compensate -- given disadvantages (e.g., \textit{Commission v France}, \textit{Griesmar}) \citep[][para. 10-11]{C-312/86}, \citep{C-366/99}, \citep{xenidis2018positive}. All in all, positive action has largely been interpreted as `an exception to' rather than `an integral part of' EU anti-discrimination law \citep[][p. 28]{mccrudden2019gender}.\footnote{Both the EU Commission and the EU Parliament have called for a wider use of positive action measures \citep{mccrudden2019gender}.} While the Directives allow for positive action in relation to all six protected grounds, the case law of the CJEU has focused on gender equality.

\section{The Lawfulness of Algorithmic Positive Action}
\label{sec:algorithmicpositiveaction}
While an emerging body of work has explored the lawfulness of algorithmic affirmative action under US anti-discrimination law \citep{bent2019algorithmic,hellman2020measuring,ho2020affirmative,kim2022race}, legal scholars have only recently started addressing algorithmic positive action under EU non-discrimination law \citep{hacker2018teaching,xenidis2022algorithmic, muller2023fairness}. In this section, we explore when such `algorithmic positive action' is likely to be lawful and when it is less likely to survive legal scrutiny. 

First, the scope of such provisions varies considerably at the national level because the EU equality Directives only permit Member States to adopt positive action measures. Some national legislation requires public as well as private actors to take positive action whereas others leave out the private sector. The capacity of public, but especially private, actors to engage in lawful fair-ml interventions to avoid discrimination or perform positive action varies therefore across the EU and might be restricted. 

While the notion of positive action covers a broad array of measures, fair-ml interventions have often been compared to quotas. Since the CJEU has defined their lawfulness strictly, however, this analogy has to be examined carefully. Some forms of fair-ml interventions proceed from awareness of existing inequalities but do not directly attribute preferences to groups or individuals.\footnote{See Section~\ref{sec:neutralityfallacy} below. In the US context, Kim argues that approaches addressing colourblindness do not automatically fall under the scope of affirmative action \citep{kim2022race}.} Fair-ml interventions that do attribute preferences, however, proceed from a concern for equality of outcomes. By contrast, the Court has focused on equality of process when assessing the lawfulness of quotas, including the justification of factors used in decision-making (e.g., are the decision criteria relevant for the outcome?) and the specific implementation of positive action measures (e.g., is the measure flexible, exceptional, temporary, and transparent?). This difference in rationales points to the broader issue of translation. How would the Court assess the formalisation of qualifications and quantification of `equal qualification'? Statistical models allow for a `precise' assessment of the suitability of two candidates but at the same time conceal the inherent uncertainty of such scores. Hence, it might be difficult for judges to assess whether a given score range or threshold effectively signals `equal qualifications'. Concretely, if a model attributes a score of 0.9 to a female candidate while her male competitor scores 0.92, would hiring the female candidate be deemed positive action? The seemingly objective nature of mathematical formalisation could make it more difficult to assess the lawfulness of positive action.  

The lawfulness of fair-ml interventions to avoid discrimination or perform positive action depends on the nature of that intervention. Some authors have noted that group fairness metrics closely resemble quota systems \citep{hacker2018teaching,muller2023fairness}. Indeed, when base rates are unequal between groups, fairness constraints such as \textit{demographic parity} (which requires equal selection rates amongst groups), or \textit{equalized odds} (which requires equal misclassification rates) require differential treatment of individuals with similar risk profiles. For example, if the proportion of qualified men in an applicant pool is higher than the proportion of qualified women, a well-calibrated and accurate resume selection algorithm will produce (on average) lower scores for women than for men. If the algorithm is unable to perfectly separate qualified from unqualified individuals, the differences in the distribution of scores between men and women will result in different distributions of false positives and false negatives across genders. In particular, unqualified men will be more likely to be misclassified as qualified compared to unqualified women. Vice versa, qualified women will be misclassified as unqualified at higher rates compared to qualified men. In other words, in this scenario, equal treatment of men and women with similar risk profiles cannot satisfy equalized odds. Similarly, if base rates are unequal across protected groups, enforcing equal selection rates implies (intentional) misclassification of some groups over others. If classification is connected to receiving a particular benefit, misclassification could therefore imply that preference is given to members of some groups over others.

The similarity of fairness interventions to quotas is most striking when we consider group-specific decision thresholds \citep[e.g.,][]{hardt2016equality}. For example, to achieve equal selection rates among female and male applicants, an employer could decide to lower the decision threshold for female applicants. As a result, a man and woman with the same score would be classified differently. Similarly, group-specific decision thresholds can be used to adjust the balance of false positives and false negatives in order to achieve equalized odds. Beyond mingling with decision thresholds, \textit{any} fairness intervention that optimizes for demographic parity or equalized odds under disparate base rates must result in skewed classifications, resulting in `hidden' quotas.

The Court's case law requires that qualifications and merit must remain the leading decision-making factors and that preferential treatment can only act as a tie-breaker. Consequently, fair-ml interventions that optimize for group fairness metrics that require differential treatment are unlikely to pass legal scrutiny (e.g., \textit{Abrahamsson} \citep[]{C-407/98}). Assuming that scores produced by a hiring algorithm are an adequate measure of qualifications and merits, some form of group-specific thresholding that acts as a tie-breaker may still be allowed. In particular, similar to the reject option classification approach proposed by \citet{kamiran2012decision}, we may devise a quota that considers two candidates equally qualified if their scores fall within a particular range (e.g., a maximum difference of 0.05). Within the pool of `equally' qualified candidates, preference could then be given to the underrepresented group.

Yet such an algorithmic intervention would need to satisfy a second criterion to ensure lawfulness. As explained above, quotas must include a saving clause that ensures that the specific circumstances of individuals is taken into consideration and can tilt the balance in their favour. This implies, first, that automatic fairness interventions in algorithmic decision-making are likely to be struck down.\footnote{It follows from GDPR art. 22 that "automatic decision making which "produces legal effects concerning him or her or similarly significantly affects him or her" (Art. 22(1)) is only allowed on certain conditions, for example explicit consent or explicit legal basis in domestic law (Art. 22(2)). In a recent ruling on credit scoring (Case C‑634/21 (Schufa)), the CJEU established that "decision making" not only refers to the final decision in a case, but also encompasses automated decisions that are made before the final decision, when those decisions play an important role in the final decision. In Schufa, an automated credit score played an important role in the banks decision on allowing or refusing a loan. Hence the credit score was considered to fall under Art. 22, thereby activating the need to fulfill one of the conditions in Art. 22(2).}
Second, as also noted by \citet{hacker2018teaching}, any human oversight would have to prove effective in reviewing attributed preferences\footnote{The notion of human-in-the-loop has been termed an `impossible figure', see \citep[]{amoore2020cloud}}. Finally, a measure must have as its objective to reduce inequalities for historically underrepresented groups, which has to be demonstrated and linked to the existing pool of applicants~ \citep{nowak2022dutch} and justified by the level of underrepresentation~\citep{wojcik2023assessing}.

In conclusion, only a very limited set of algorithmic fairness interventions are likely to be interpreted as lawful measures of positive action. Specifically, this includes measures that map onto the conditions for quotas set by the Court: preferential treatment can only act as a tie-breaker, must include a saving clause, and must be proportional.

\section{The Neutrality Fallacy: When Fairness Interventions are Not Positive Action}
\label{sec:neutralityfallacy}
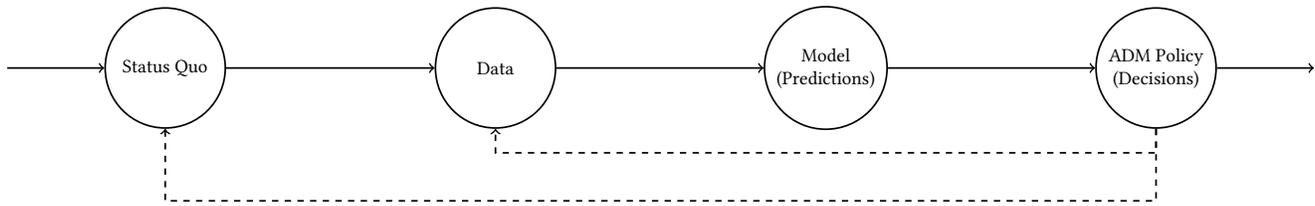
\begin{figure*}[ht]
    \centering
    \def\sp{5.5}
    \resizebox{\textwidth}{!}{%
        \begin{tikzpicture}[node/.style={draw, align=center, circle, minimum width=20mm}, thick,align=center]
        \node (history) at (0,0) {};
        \node [node] (sq) at (0.5*\sp,0) {Status Quo};
        \node [node] (data) at (1.5*\sp,0) {Data};
        \node [node] (model) at (2.5*\sp,0) {Model\\(Predictions)};
        \node [node] (dec) at (3.5*\sp,0) {ADM Policy\\(Decisions)};
        \node (future) at (4*\sp,0) {};
        \draw[->] (history) -- (sq) node[above, midway] {};
        \draw[->] (sq) -- (data) node[above, midway] {};
        \draw[->] (data) -- (model) node[above, midway] {};
        \draw[->] (model) -- (dec) node[above, midway] {};
        \draw[->] (dec) -- (future) node[above, midway] {};
        \draw[->, dashed] (dec.south) |- node[below,xshift=-155]{} ($(dec.south) - (0,4mm)$) -|  (data.south);
        \draw[->, dashed] (dec.south) |- node[below,xshift=-240]{} ($(dec.south) - (0,12mm)$) -|  (sq.south);
        \end{tikzpicture}
    }
    \caption{Neutrality fallacies occur when an artefact of algorithmic decision-making is assumed to be "neutral".}
    \label{fig:neutralityfallacy}
\end{figure*}

As the CJEU has been strict in assessing positive action measures, interpreting fair-ml interventions as positive action would impose a significant legal burden on those wishing to implement them. However, categorizing a fair-ml intervention as positive action relies on very strong assumptions regarding the neutrality and objectivity of algorithmic decision-making.\footnote{See also \citet{friedman1996bias}.} In the remainder of this section, we identify three instances of this category mistake, which we refer to as \textit{neutrality fallacies}. We suggest that neutrality fallacies occur when artefacts in different stages of algorithmic decision-making are assumed to be "neutral" in relation to the \textit{status quo}, which challenges the relevance of designating fair-ml measures as positive action. This includes the data that is used to train a model, the model that produces predictions, and the algorithmic decision-making policy that prescribes how predictions are acted upon (Figure~\ref{fig:neutralityfallacy}).

\subsection{"Data is Neutral"}
Considering any fair-ml intervention as a form of positive action must rely on the assumption that the data that is used to train the model is `neutral' towards a given social reality. This is not always a reasonable assumption.

The goal of predictive modelling is to accurately predict some outcome of interest. In real-world decision-making settings, the outcomes of interest are often unobservable constructs: complex phenomena that do not have an immediate physical representation. For example, we may be interested in predicting employee qualifications, creditworthiness, recidivism risk, or healthcare needs. While we have a theoretical understanding of these constructs, they cannot be measured directly. Instead, they must be inferred from measurements of other, observable properties. This requires making assumptions about the relationship between measurable properties and the outcome of interest. For example, measuring a person's socio-economic status by their income relies on the assumption that income meaningfully relates to our theoretical understanding of socio-economic status \citep{jacobs2021measurement}.

In machine learning, the outcome of interest is operationalized by the `ground truth' target variable. For example, in credit risk assessment, the construct creditworthiness is often operationalized by loan defaults. More often than not, target variables are simple proxy variables for the outcome of interest that are readily available. This makes machine learning datasets vulnerable to measurement bias: a systematic difference between the target outcome of interest and its operationalisation by the target variable \citep{guerdan2023ground}. For example, while income is certainly related to socio-economic status, it does not capture other relevant aspects such as capital ownership, education or wealth. Measurement bias is problematic because a predictive model that predicts a target variable, which do not adequately measure the outcome of interest is less likely to meaningfully support decision-making. 

There are many ways in which measurement bias can materialize. First of all, using a single proxy attribute to capture a complex phenomenon is unlikely to fully capture all relevant aspects of the outcome of interest. Another important source of measurement bias is a mismatch between the collected data and the envisioned deployment context. In some cases, the data sample population differs from the population in the envisioned deployment context. For example, patient insurance data that was collected in a French hospital is likely not representative of patient populations in other countries. In other cases, the collected data may reflect a historical decision-making policy subject to change. For example, a dataset that captures job seekers' prospects on the job market necessarily reflects their chances subject to the support they received from the government. If a government's resource allocation policy changes, the dataset may no longer capture job seekers' prospects. Additionally, data may not be collected at random, which paints a distorted picture. For example, fraud analysts at a bank are typically not able to manually inspect all transactions for fraud. Instead, they may rely on hand-crafted rules or automated fraud detection algorithms to determine which transactions should be inspected. As a result, observed fraud is only a subset of true fraud, skewing the dataset towards fraud cases that match the analysts' expectations.

When measurement bias is associated with protected group membership, it can become a source of discrimination \citep{jacobs2021measurement}. 
The most obvious instance of this phenomenon occurs when target labels encode implicit bias present in historical decision-making. For example, a recruiter may be less likely to attribute qualities that are usually stereotypically conceived of as `male', such as ambition and self-confidence, to a female candidate \citep{xenidis2019eu}. A female applicant's sex may subconsciously play a part in a recruiter's mental decision-making process, negatively affecting their overall impression of the applicant \citep{adams2022regulating}. As a result, past employment decisions may be a biased measurement of an applicant's true qualifications. A predictive model that is trained to predict past employment decisions will reproduce the implicit bias encoded in the labels.
Implicit bias can also rear its head in more subtle ways when it affects which data is (not) collected. For example, arrest records are an imperfect proxy for recidivism risk. Not only do not all arrests correspond to a crime (a form of label bias similar to our hiring example), but also not all crimes are captured in arrest records. When arrests are affected by racist policing practices, measurement bias is higher for racial groups that are most affected by policing. When the actual outcome of interest is recidivism risk, the quality of decisions based on re-arrest predictions will therefore be worse for already disadvantaged groups. 
Measurement bias can also reflect tangent structural inequalities. For example, in contexts where access to healthcare is affected by structural discrimination, healthcare costs will be a biased proxy for healthcare needs. When a decision-support tool trained to predict healthcare costs is used to allocate resources, structural inequalities in access to care are reproduced \citep{Obermeyer2019}.

When a target variable is affected by measurement bias, inequalities do not arise from a person's materialized abilities or characteristics, but from our erroneous beliefs about their abilities or characteristics \citep{weerts2022does}. The legal position should be clear: if decisions affected by implicit bias of decision-makers are discriminatory, reproducing this bias in a predictive model is also discriminatory.\footnote{\citet{adams2023directly} note that the UK Supreme Court has held that implicit bias could even fall into the scope of direct discrimination (`subjective direct discrimination'). Referring to the decision in \textit{CHEZ}, Advocate General Kokott explained that the CJEU 'considers a measure taken on the basis of stereotypes and prejudices in relation to a particular group of individuals to be an indication of direct discrimination' \citep[][fn. 30]{C-157/15}.} For example, accounting for prejudice entrenched in a target variable requires some consciousness of gender, but a failure to intervene (which we could call `gender blindness') would give men an unfair advantage. Even though measures aimed at addressing measurement bias require an (implicit) awareness of protected group membership, they should not be considered a violation of the principle of equal treatment.\footnote{\citet{kim2022race} makes a similar argument in relation to affirmative action under US anti-discrimination law.} Instead, such interventions are best construed as measures to avoid discrimination\footnote{\citet{wachter2020bias} even go as far as to argue that in domains such as employment, where discrimination is well-documented and pervasive, an absence of interventions should be sufficient to raise \textit{prima facie} discrimination.}, rather than measures of positive action.

The legal question then becomes whether measurement bias has occurred and to what extent a fairness intervention is an effective and justifiable measure to avoid discrimination arising from that bias. It can be difficult to disentangle materialized structural inequalities from automated stereotyping and prejudice. When we observe an association between a protected characteristic and a target variable, we cannot tell from the data itself whether and to what extent the target variable is affected by measurement bias or accurately reflects the (unequal) social reality. From a moral and legal perspective, however, the difference is important. While mitigating measurement bias is closely related to a notion of formal equality, addressing inequalities in the status quo promotes substantive equality \citep{weerts2022does}. Compared to disparities caused by measurement bias, interventions to address disparities caused by an unequal status quo open up more possibilities for objective justifications under the indirect discrimination doctrine and, in some cases, could even be considered positive action.

The assessment of measurement bias in the target variable requires convincing empirical assumptions. For example, if racial disparities in access to healthcare in the deployment context have been documented in scientific research, these studies could substantiate the argument that past healthcare costs do not adequately reflect healthcare needs. Contrarily, if predictions for healthcare costs in a deployment context correlate well with established measurements for healthcare needs, such as the number of active chronic conditions, this could strengthen our belief that healthcare costs are a reasonable proxy for healthcare costs.\footnote{\citet{Obermeyer2019} show the exact opposite trend considering race in their analysis of a commercial predictive algorithm that is widely used in the United States.} We refer to \citet{jacobs2021measurement} for an overview of how such assumptions can be tested in practice.

When measurement bias causes an association between a protected characteristic and a target variable, a machine learning model is likely to replicate this association. When the protected characteristic is not available in the data set, the model could still replicate the association through proxies of the protected characteristic, resulting in a form of proxy discrimination. A way to mitigate this problem is to remove (obvious) proxies for protected characteristics from the data set, but this strategy suffers from at least two flaws. First, machine learning algorithms are specifically designed to identify relationships between features and the target variable. If there is a relationship, it is likely reproduced through more subtle paths. For example, \citet{davies2022learning} identify a set of variables that together fully capture the correlation between a protected characteristic and the target variable in a criminal justice setting. Second, removing features that are associated with a protected characteristic could harm the performance of the predictive model.

Another potential intervention is to use one of the classical fairness-aware mitigation algorithms that optimizes for equal selection rates between groups (i.e., demographic parity). Yet this intervention is likely to be ineffective because fair-ml mitigation algorithms generally do not encode explicit assumptions on how measurement bias materializes in a dataset. As a result, there is no guarantee that predictions that satisfy demographic parity correspond to a better representation of reality than the target variable in the training data \citep{weerts2022does}. For example, without making assumptions about the measurement process, we cannot tell exactly which people are more likely to have been affected by the measurement bias. Additionally, equalizing selection rates between groups relies on the assumption that measurement bias is the only factor responsible for differences in base rates. However, in domains where this type of measurement bias is most likely to occur, structural inequalities are typically pervasive as well. In the absence of empirical assumptions, it is therefore impossible to tell where to draw the line. The algorithm could overcorrect and enter the domain of positive action.

More promising are interventions that explicitly target measurement bias. Tackling measurement bias is never a matter of simply collecting \textit{more} data. Instead, we need to collect \textit{other} data that enables a better representation of the construct we are interested in. The most straightforward intervention is to choose a different target variable that does not suffer from measurement bias. When measurement bias is caused by non-random selection of instances, a different data collection procedure could be used. For example, \citet{weerts2023look} suggest the use of active learning to mitigate selection bias in scenarios where random selection is infeasible. In cases where it is not possible to collect better measurements, it might still be possible to carefully model measurement bias, quantitatively formalizing the assumed relationships between the available data and the unobserved outcome of interest. For example, \citet{guerdan2023counterfactual} propose an empirical risk minimization method that, given knowledge of the properties of measurement bias, corrects for it.

\subsection{"Predictive Models are Neutral"}
Even when the target variable is a reasonable proxy for an outcome of interest, to what extent can a predictive model accurately predict it? A model may not produce equally accurate predictions across protected groups.
When a predictive model falls within the material scope of EU discrimination law, disparities in predictive performance could constitute discrimination. Case law in this area is scarce, but several scholars have argued that EU non-discrimination law could be applicable \citep{caraccioloditorella2022,weerts2023algorithmic,adams2023directly}. If that is the case, measures taken to mitigate performance disparities are best considered as measures to avoid discrimination, rather than positive action.
For example, the Netherlands Institute for Human Rights recently judged that the dating app Breeze must take measures to ensure that their algorithm does not amplify users' preferences in such a way that the matching probability is less accurate for dark-skinned or non-Dutch users \citep{breeze2023}. The question then becomes whether an intervention is effective and does not have (discriminatory) side effects. 

The effectiveness of an intervention depends on the source of the performance disparity.
A machine learning model is typically optimized to minimize errors over the entire training data set. If the data distribution of a minority group differs substantially from the majority and the model class is not sufficiently complex, the model could underfit the minority group, resulting in performance disparities. Various legal scholars have argued that if being aware of a protected attribute helps to predict an outcome of interest, it should not necessarily be considered a violation of equal treatment to take that attribute into account \citep{ho2020affirmative,hellman2020measuring,hoffman2022fairness,kim2022race}. For example, it is known that heart disease can present differently in female patients than in male patients~\citep{maas2010gender}. Supplying a patient's sex as a feature in a clinical prediction model could help the model distinguish between individuals who are more or less likely to have particular symptoms, which could improve predictive performance.

\citet{kim2022race} argues that models that use protected group membership during training but not inference may pass legal scrutiny. In particular, a protected characteristic could be used (implicitly) in the optimization objective of the machine learning algorithm. For example, oversampling a minority group in the training data penalizes the model for misclassifying members of this group, which could lead to an increase in predictive performance for that group. \citet{kim2022race} argues that such measures should not raise legal concerns, as "these types of strategies are more accurately understood as removing bias from processes that would otherwise be unfair". Moreover, one of the conditions for discrimination to occur is that treatment is `less favourable' or a person experiences a `particular disadvantage'. Measures such as oversampling will not necessarily create a \emph{disadvantage} for non-protected groups: it may very well be possible to identify a predictive model that is equally accurate for the non-protected group, yet more accurate for the protected group. In such cases, no real preference is given, and such a model would not necessarily warrant strict judicial review. However, it must be noted that there could be unwanted side effects. For example, if base rates differ between protected groups, group-specific sampling schemes skew the base rates in the training data set. Unaccounted for, over- or undersampling can therefore result in miscalibration of risk scores, which could skew subsequent decision-making.\footnote{For example, we may oversample female patients during the training of a cancer risk prediction model in the hope that it increases the model's accuracy for women. If this particular type of cancer occurs more often in women compared to men, oversampling women increases the prevalence of patients with cancer in the training dataset. As a result, a machine learning model trained on the oversampled dataset is likely to produce higher risk scores overall -- corresponding to the increased observed cancer prevalence among patients in the dataset as a whole -- than that suggested by the original dataset before changing the sampling. Unaccounted for, these risk scores cannot be interpreted as probabilities and could skew subsequent (automated) decision-making.}

Even including a protected characteristic explicitly as a feature in a machine learning model could be lawful in some cases. \citet{hellman2020measuring} argues that if a particular feature is a good predictor for group $A$ but not group $B$, excluding the criterion for classification of group $B$ would not constitute disparate treatment. For example, we may train a decision tree that splits into separate branches based on a protected characteristic. Again, if the use of a protected class variable in a machine learning model does \textit{not} result in \textit{harm}, differential treatment does not amount to discrimination. Here the differential treatment would not harm group $B$.
In contrast, using a protected characteristic to directly increase or decrease a risk score is less likely to pass legal scrutiny. For example, a linear regression model with a positive coefficient for a protected characteristic essentially acts like a mechanical thumb that increases the predicted score for specific groups. In the absence of a causal connection between the protected characteristic and the target, this would almost certainly be viewed as a violation of the principle of equal treatment -- even if it results in more accurate risk scores. For example, in \textit{Test-Achats}, the Court judged that the use of gender in the determination of insurance premiums as a proxy for other distinguishing factors such as life expectancy or driving behaviour is incompatible with the principle of equal treatment \citep{C-236/09}. 

The appropriateness of the inclusion of a protected characteristic also depends on how the protected characteristic is operationalized. Several scholars have underlined the importance of ensuring appropriate operationalization of protected characteristics that correspond to complex social constructs, such as race \citep{hanna2020towards,wojcik2023assessing}. For instance, while self-identified race is a known risk factor in first-episode psychosis \citep{kirkbride2017ethnic}, measuring race as a phenotypical presentation does not capture the individual discrimination experience that appears to be at the root of the issue. Including phenotype rather than self-identified race as a feature in a clinical risk prediction model for psychosis could therefore be problematic. Additionally, the inclusion of protected characteristics can inadvertently reinforce stereotypes. For example, medical professionals often misinterpret a correlation between race and clinical outcomes for biological causation rather than social causation, a belief that is reinforced by race-corrected clinical prediction models \citep{wojcik2023assessing}.

\subsection{``Algorithmic Decision-Making is Neutral"}
Predictive models are often presented as methods that can assist decision-makers in `objective' decision-making.  For example, \citet{kwegyir2023misuse} note that the developers of PATTERN, a criminal risk assessment tool, state that it ``is a neutral assessment tool, as evidenced by the nearly equal scores [for different racial groups] on the Area Under the Curve (AUC) analysis". However, even if predictions are valid and accurate, the assumed `neutrality' cannot be extended to the decisions based upon them.

Any decision-making policy, be it human or algorithmic, must weigh different consequences of a decision based on the available evidence. The value-laden nature of a decision-making policy is the most obvious in automated decision-making, which requires explicitly delineating how a prediction will be translated into a particular decision or action. Perhaps the simplest decision policy is the use of a decision threshold in binary classification scenarios. When risk scores are calibrated, the decision threshold explicitly encodes how much we value false positives over false negatives: the higher the threshold, the costlier we deem false positives relative to false negatives. For example, the Public Employment Service of Austria (AMS) used a machine learning model to predict job seekers' employment prospects \citep{allhutter2020algorithmic}. The allocation policy is straightforward: job seekers with the highest prospects (a probability of $>0.66$ to find employment for at least 3 months within the next 7 months) and lowest prospects (a probability of $<0.25$ to find employment for at least 6 months within the next 2 years) receive fewer resources, while job seekers who do not belong to either of these categories will receive additional support measures. The AMS claimed this to be an `objective' and `neutral' profiling system, that simply reflects the `harsh realities' of the job market \citep{allhutter2020algorithmic}. However, not providing resources to job seekers with a high risk of long-term unemployment explicitly encodes neoliberal values, which is a normative choice. By not supporting those with the lowest prospect of employment, this group will continue to struggle. In contrast, other countries such as Sweden use statistical profiling to support additional support to job seekers with the lowest prospects, thereby improving the chances that this group will find relevant jobs \citep{scoppetta2018tackling}.

More complex decision-making policies can be determined through quantitative optimization. Take for example the use of no-show risk predictions in medical appointment scheduling. Scheduling policies often overbook particular slots to minimize schedule cost, which is quantified as the weighted sum of total patient waiting time and the provider's overtime and idle time. \citet{samorani2022overbooked} show that patients belonging to groups characterized by a higher no-show risk are disproportionately scheduled into overbooked slots, resulting in systematically longer waiting times for these patients. Moreover, they show that an alternative \textit{min-max} optimization objective, which minimizes the waiting time of the group expected to wait the longest, can mitigate disparities at a similar scheduling cost to the original objective. In other words, even when patients' no-show risk estimates are accurate and valid, the decision-making policy that informs how we act upon the predictions encodes specific value judgements.

While some algorithmic decision-making policies could amount to positive action (e.g., when the policy explicitly gives preference to some group over others), merely replicating the status quo cannot be thought of as `neutral' and taking steps to consider the effect of different decision-making policies on protected groups would fall within the scope of prevention against discrimination.
\\ \\ 
To summarize, in this section, we have argued that fairness interventions of different kinds should not be considered through the strict judicial review lens usually applied to positive action measures if they address the neutrality fallacies in data, models and decision-making policies exposed above.

\section{Towards A Positive Obligation to Avoid Discrimination?}
\label{sec:positiveobligation}
EU discrimination law seemingly draws a clear line between the \textit{obligation} to avoid both direct and indirect discrimination and the mere \textit{possibility} to adopt positive action measures. This distinction is important because it determines the lawfulness of fair-ml interventions. Positive action is subjected to careful judicial review, especially when it takes the form of quotas. As shown by our analysis, avoiding discrimination in the context of AI requires taking active steps to avoid the replication, amplification, or even introduction of inequalities. For some fair-ml interventions, this raises the question of where they fall on the spectrum between the prevention of discrimination and positive action, and \textit{in fine} of their lawfulness.

The problem is that the delineation is not as clear-cut as it might seem: where does refraining from discrimination end and positive action start? In fact, this dividing line is artificially constructed and contingent on the baseline we choose to represent social reality (as manifested e.g., in the legal figure of the comparator). In turn, this baseline ultimately depends on an implicitly projected `ground truth’ concept that necessarily entails some form of social engineering. For example, if the AMS algorithm should not reproduce `the harsh reality' of the labour market, what ground truth should it represent? The claim to neutrality inherent in framing interventions as simply `correcting' for biases ignores the historical construction of inequalities that shape the present social reality. In other terms, taking the current social \textit{status quo} as a baseline to distinguish between non-discrimination and positive action simply amounts to another neutrality fallacy: the assumption that the status quo is neutral. Both the ban on indirect discrimination and positive action rely on the recognition that historical inequalities have been naturalised over time and have become `invisible'. Thus, they form two sides of the same coin and drawing a line between both for purposes of preventing social engineering may not be meaningful. It might even deter users of algorithmic decision-making systems from attempting to engage in fairness work for fear of liability.  

Ultimately, fair-ml interventions raise the question of whether a negative obligation to refrain from discrimination is sufficient and appropriate in the context of algorithmic decision-making. Given the scale of algorithmic bias in certain sectors, we ask whether the obligation not to discriminate should not be recast as a positive duty to take proactive and meaningful action to prevent unlawful discrimination. We suggest moving away from a \textit{negative} obligation to \textit{"not do harm"} towards a \textit{positive} obligation to actively \textit{"do no harm"} as a more adequate framework for algorithmic decision-making and fair ml-interventions.\footnote{For the avoidance of doubt, we do not suggest that fair-ml interventions should never face judicial review. Instead, our proposal emphasizes a shift from demonstrating the absence of discriminatory effects, towards demonstrating that appropriate and effective actions have been taken to ensure that unlawful discrimination is avoided. Our proposal thus aligns with recent work that calls for a duty of reasonable search for less discriminatory algorithms in United States anti-discrimination law~\citep{black2023less}.}

Moving towards a positive obligation to prevent discrimination in algorithmic processing could gain a foothold in EU law through the upcoming AI Act. While the Act has yet to be adopted, a political agreement was reached in early December in the so-called trilogue negotiations. The Council's general approach to negotiations (which is considered close to the final text that will be adopted) indicates that providers of high-risk AI systems (which include for example welfare services, employment, education and law enforcement) will have to document that these systems are not discriminatory. Risk management obligations include data governance, technical documentation, transparency and accuracy (chapter 2 of the proposed Act). This obligation on providers will transfer to developers and since providers can receive fines for breaching the Act, they will likely impose strict contractual obligations on developers to ensure compliance with the AI Act.

EU law also contains other forms of positive action that can serve as a yardstick to model such a positive obligation. For example, the obligation of employers to offer reasonable accommodation to workers living with a disability requires that active measures be taken to adapt the workplace to prevent the exclusion of those workers from the labour market. Failure to adopt such measures -- in a manner proportionate to the size of the employer as signalled by the notion of reasonableness -- amounts to discrimination. Expanding on this framework to craft a positive legal obligation to prevent algorithmic discrimination would effectively tighten the grasp of non-discrimination law onto algorithmic decision-making.

Concretely, what form could a positive obligation to prevent algorithmic discrimination take? Considering machine learning models specifically, we can draw from the neutrality fallacies identified in the previous section. Specifically, a positive obligation could introduce a duty to justify the use of a target variable that is associated with a protected characteristic, the use of a machine learning pipeline in which predictive performance disparities are explicitly taken into account, and the justification of an (automated) decision-making policy. Future work should focus on the implications of such a positive obligation in existing non-discrimination law as well as new regulatory frameworks, such as the AI Act. 

Finally, we would like to emphasise the following. Even in cases where algorithmic fairness interventions can be regarded as legitimate measures to prevent discrimination, their effectiveness will depend on the existence of additional interventions. Positive action remains indispensable to address structural discrimination holistically in society. Positive action measures have the power to truly transform the baseline on which `neutrality' is premised to break the circle of historical inequalities. As stated by Advocate General Tesauro in \textit{Kalanke}~\citep{C-450/93}: "Formal, numerical equality is an objective which may salve some consciences, but it will remain illusory and devoid of all substance unless it goes together with measures which are genuinely destined to achieve equality [...] [T]hat which is necessary above all is a substantial change in the economic, social and cultural model which is at the root of the inequalities – a change which will certainly not be brought about by numbers and dialectical battles which are now on the defensive." This view has been reflected in some policy developments in Europe. For example, the recent `Women on Boards' Directive 2022/2381~\citep{2022/2381} sets a positive obligation to improve the gender balance among directors of listed companies such that at least 40\% of directors are female. A similar move can be observed in other jurisdictions. For example, in 2022, Australia introduced a legal obligation to take `reasonable and proportionate measures' to eliminate, amongst others, sex discrimination in employment.

\section{Conclusion}
\label{sec:conclusion}
Scholars have warned that fairness interventions in algorithmic decision-making could fall within the scope of positive action in EU law, especially when they resemble quotas. This imposes a significant legal burden on AI users and providers as they fall within a double bind: a prohibition to engage in discrimination and simultaneously very strict limits on how they might redress such discrimination \textit{ex ante} to avoid liability. In this paper, we have shown that not all fairness interventions should be classified as positive action. Doing so would reflect the erroneous assumption that the data, models and policies that constitute algorithmic decision-making are neutral. We have exposed these neutrality fallacies to argue that in the context of algorithmic decision-making, active preventive steps are necessary to avoid AI reproducing, amplifying and enacting inequalities. We, therefore, suggest moving away from a \textit{negative} obligation to \textit{not do harm} towards a \textit{positive} obligation to actively \textit{do no harm} as a more adequate framework for algorithmic decision-making and fair ml-interventions. Such a positive obligation to prevent algorithmic discrimination has the power to increase public discussions around preventive measures and their effectiveness, as well as facilitate accountability among providers and users. In this framework, the justification of fairness interventions under EU law ought indeed to be precise, including technical, empirical, and normative assumptions. Importantly, our proposal should not preclude -- but instead, complement -- holistic forms of positive action addressing the root causes of inequality and aimed at breaking the circle of historical inequalities in and beyond the algorithmic society.

\section*{Research Ethics and Social Impact}

\subsection*{Ethical Considerations Statement}
This work does not describe experiments with users and/or deployed systems and does not rely on sensitive user data. 

\subsection*{Researcher Positionality Statement}
The research, disciplinary backgrounds, and personal views of the authors have influenced this work. At the time of doing this research, all authors were employed at universities in the European Union. Several of the authors draw from their work on (a) research at the intersection of European discrimination and equality law and technology (b) AI engineering experience in building AI auditing systems with a focus on fairness, (c) fair-ml research contributing novel fair-ml approaches and their validation in real worlds use cases and interactions with practitioners.

\subsection*{Adverse Impact Statement}
There are some ways in which our work, once published, could have an adverse impact. Specifically, our arguments could be misinterpreted, misused, or misconstrued as carte blanche to apply algorithmic fairness interventions. Instead, our work is intended to expose the value-laden nature of design choices in algorithmic decision-making, including but not limited to algorithmic fairness interventions.

\begin{acks}
This project has received financial support from the CNRS through the MITI interdisciplinary programs through its exploratory research program. We would like to thank the Lorentz Center and its support team as well as the organizers and participants of the workshop \textit{Fairness in Algorithmic Decision Making: A Domain-Specific Approach} for the stimulating discussions which brought about our interdisciplinary collaboration.
\end{acks}

\bibliographystyle{ACM-Reference-Format}
\bibliography{references}

\end{document}